\documentclass[10pt,letterpaper]{article}
\usepackage{cogsci}
\cogscifinalcopy 
\usepackage[hidelinks]{hyperref}
\usepackage{graphicx}
\usepackage{xcolor}
\usepackage{pslatex}
\usepackage{apacite}
\usepackage{float} 
\hypersetup{colorlinks, linkcolor={red!50!black}, citecolor={blue!50!black}, urlcolor={blue!80!black}}

\title{Self-supervised learning of video representations from a child's perspective}

\author{
  {\large \bf Emin Orhan$^{\dag}$}\ \ \ \ \ \ \   
  {\large \bf Wentao Wang$^1$} \ \ \ \ \ \ \ 
  {\large \bf Alex N. Wang$^2$} \ \ \ \ \ \ \ 
  {\large \bf Mengye Ren$^{1,2}$} \ \ \ \ \ \ \ 
 {\large \bf Brenden M. Lake$^{1,3}$} \\
 \vspace{1mm} \\
 $^{\dag}$Correspondence: {aeminorhan@gmail.com}
 \vspace{1mm} \\
 $^1$Center for Data Science, $^2$Department of Computer Science, $^3$Department of Psychology, New York University
}

\begin{document}

\maketitle

\begin{abstract}
Children learn powerful internal models of the world around them from a few years of egocentric visual experience. Can such internal models be learned from a child's visual experience with highly generic learning algorithms or do they require strong inductive biases? Recent advances in collecting large-scale, longitudinal, developmentally realistic video datasets and generic self-supervised learning (SSL) algorithms are allowing us to begin to tackle this \textit{nature vs.~nurture} question. However, existing work typically focuses on image-based SSL algorithms and visual capabilities that can be learned from static images (\textit{e.g.}~object recognition), thus ignoring temporal aspects of the world. To close this gap, here we train self-supervised \textit{video models} on longitudinal, egocentric headcam recordings collected from a child over a two year period in their early development (6-31 months). The resulting models are highly effective at facilitating the learning of \textit{action concepts} from a small number of labeled examples; they have favorable data size scaling properties; and they display emergent video interpolation capabilities. Video models also learn more accurate and more robust \textit{object representations} than image-based models trained with the exact same data. These results suggest that important \textit{temporal} aspects of a child's internal model of the world may be learnable from their visual experience using highly generic learning algorithms and without strong inductive biases.

\end{abstract}

\section{Introduction}
Children develop sophisticated visual models of the world early in their development. Whether this feat requires strong innate inductive biases or whether it can be achieved simply by applying highly generic but scalable learning algorithms to a child's visual experience is arguably one of the most significant open questions in developmental psychology \cite{elman1996,spelke1994}. 

Recent advances in our ability to collect large-scale, longitudinal video datasets recorded from the perspective of young children over the course of their early development \cite{sullivan2021} and the development of highly effective generic self-supervised learning (SSL) algorithms in machine learning \cite{caron2021,he2022} are allowing us to finally begin to tackle this modern version of the age-old \textit{nature vs.~nurture} question \cite{Wood2024}. Several recent works have already taken advantage of these advances to try to understand what kinds of visual capabilities can be learned from large-scale, developmentally realistic video data using highly generic, state-of-the-art SSL algorithms and without assuming strong inductive biases \cite{bambach2018,orhan2020,Orhan2023NMI,zhuang2021,zhuang2022}. These works typically use image-based SSL algorithms and, as a result, only consider visual capabilities that can be learned from static images, such as object recognition. However, the visual world is intrinsically temporal and important aspects of it can only be learned if this temporal dimension is taken into account. For example, the acquisition of \textit{action concepts} or the ability to predict the changes unfolding in a visual scene both require temporal information.

Here, we address this shortcoming by training self-supervised \textit{video models} on a large-scale, longitudinal, developmentally realistic video dataset, namely SAYCam \cite{sullivan2021}. We evaluate the capabilities of the trained models on several downstream tasks, compare them against a number of reference models, and provide both qualitative and quantitative insights into the learned video representations. Code and models are available from the following repository: \href{https://github.com/eminorhan/video-models}{https://github.com/eminorhan/video-models}.

\section{Methods}
\subsection{Training data}
\subsubsection{SAYCam.} 
We use the SAYCam dataset as a realistic proxy of the visual experiences of a developing child \cite{sullivan2021}. SAYCam is a large, longitudinal audiovisual dataset of headcam recordings collected from three young children (S, A, and Y) during the course of their early development (6-31 months). It contains 194, 141, and 137 hours of video from S, A, and Y, respectively, for a total of 472 hours of video. The data from each child typically consists of 1-2 hours of continuous, natural, uninstructed recordings per week. We train models on the combined data from all three children (denoted SAY below), as well as on data from child S only. 

\subsubsection{Kinetics-700.} To investigate the effect of training data on the model behavior and performance, we also train models on the full Kinetics-700 dataset \cite{smaira2020} and a randomly selected 200-hour subset of it (denoted Kinetics-200h below). The latter contains roughly the same amount of video data as child S in SAYCam and is intended as a length-matched control for child S to isolate the effect of data type alone. Kinetics-700 is a large and diverse dataset of short YouTube clips representing 700 fine-grained action categories such as \textit{playing poker}, \textit{polishing furniture}, \textit{cutting cake}, \textit{ironing hair}, \textit{etc.} The Kinetics-700 training set contains 536K video clips, each clip typically lasting shorter than 10 seconds, for a total of 1330 hours of video. Thus, compared to SAYCam, the videos in Kinetics-700 are much more diverse in content and have a much shorter time scale.

\begin{figure}
  \centering
    \includegraphics[width=0.49\textwidth, clip]{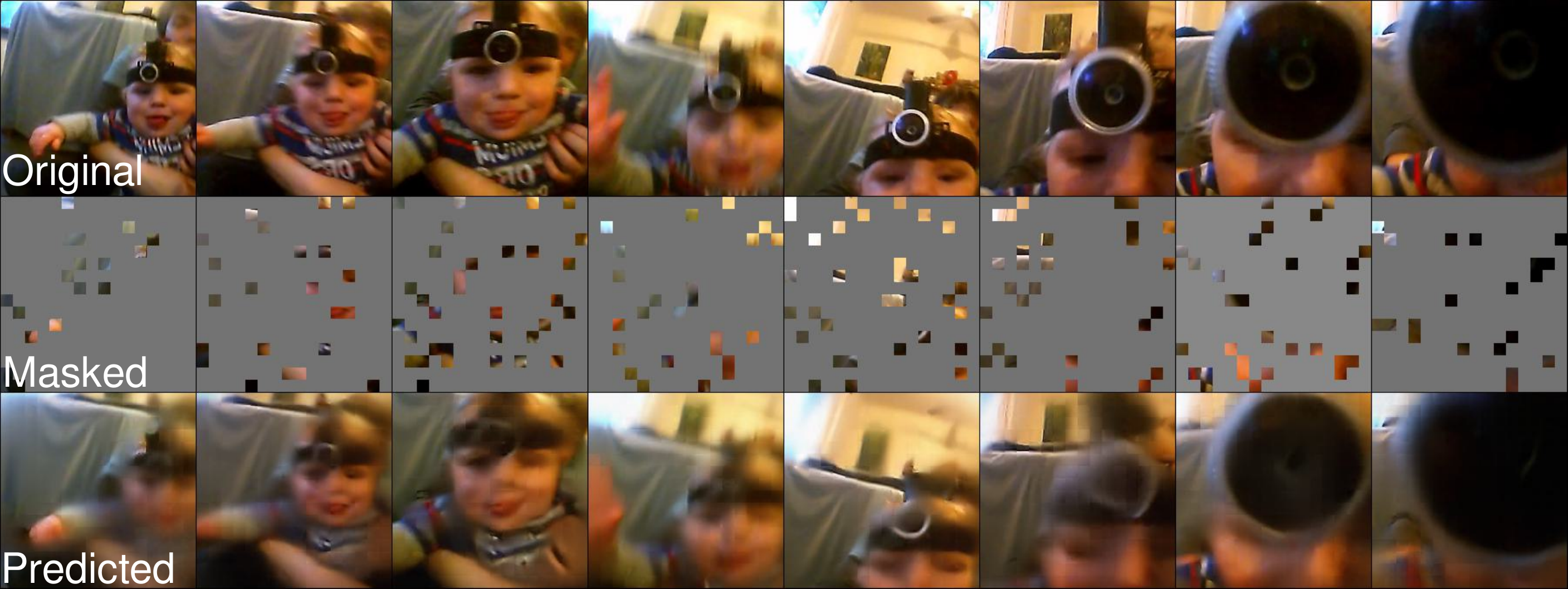}
    \vspace{-5mm}
  \caption{Illustration of the spatiotemporal MAE objective. Top row shows the original sequence of frames (from child S in SAYCam). Middle row shows the masked sequence, where 90\% of the spatiotemporal ``patches'' are randomly masked out. Bottom row shows the predictions from a model  trained on child S. The model is trained to predict the masked patches from the visible patches at the pixel level.}
  \label{maest_fig}
\end{figure}

\begin{figure}
  \centering
    \includegraphics[width=0.49\textwidth, clip]{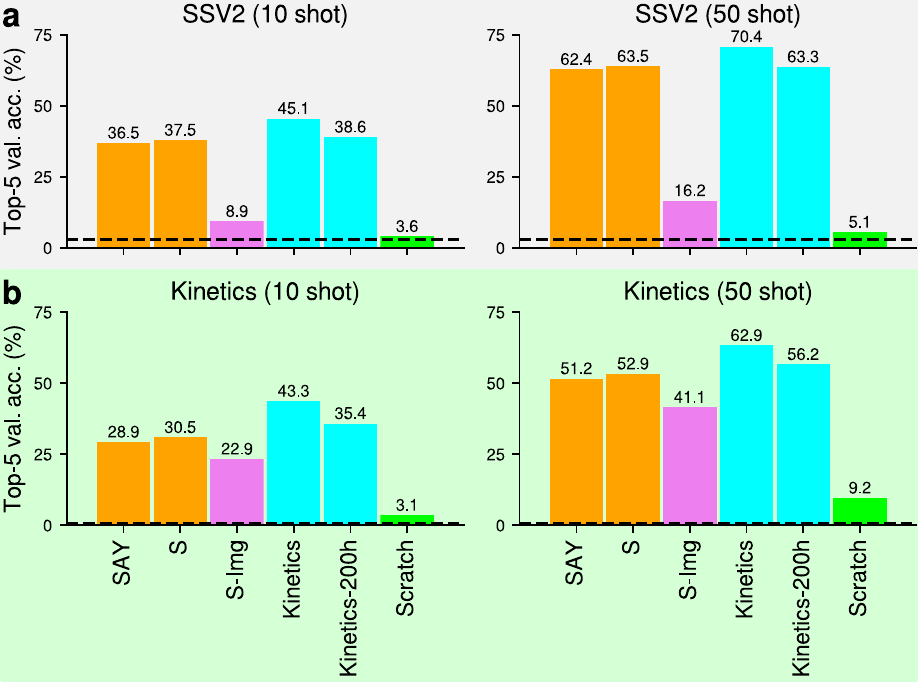}
    \vspace{-5mm}
  \caption{Top-5 validation accuracy in the SSV2 (a) and Kinetics-700 (b) action recognition tasks.~Results are shown for both 10-shot (left) and 50-shot conditions (right).~Dashed horizontal lines show the chance baseline.~{\color{orange}Orange} represents the models pretrained with the child headcam data.~{\color{cyan}Cyan} represents the models pretrained with Kinetics-700 data.~{\color{magenta}Magenta} represents a purely image-based reference model.~{\color{green}Green} represents a reference model trained from scratch on the downstream task only (no pretraining).}
  \label{accuracies_fig}
\end{figure}

\subsection{Model architecture}
We use vision transformers (ViTs) as our model architecture \cite{dosovitskiy2020}. This choice is effectively dictated by our choice of SSL algorithm, as described below. We use a large 633M parameter model (ViT-H/14) in all our experiments. We temporally subsample the videos at a rate of 3.75 frames/second and feed the model input clips consisting of 16 consecutive frames with a spatial resolution of 224$\times$224 pixels. Each modeled clip is thus roughly 4.3 seconds long. These clips are divided into 2$\times$14$\times$14 three-dimensional boxes or ``patches'' (\textit{i.e.}~2 frames in the temporal dimension, 14$\times$14 pixels in the spatial dimensions). The patches are then linearly projected onto a common patch embedding space and separable (and learnable) spatial and temporal position embeddings are added to the patch embedding of each patch, helping to identify its spatial and temporal position. The rest of the architecture is a standard transformer model operating on the flattened patch representations. Since this is a standard architecture, we refer the reader to \citeA{feichtenhofer2022} for further details.

\subsection{SSL algorithm}
We use spatiotemporal masked autoencoders (MAEs) as our SSL algorithm of choice \cite{feichtenhofer2022}, which is a straightforward extension of the image-based MAEs \cite{he2022} to video data. The basic idea in spatiotemporal MAEs is to randomly mask out a large proportion (90\%) of the three-dimensional ``patches'' described above and to predict these masked patches from high-level representations of the visible patches (Figure~\ref{maest_fig}). An MAE consists of an encoder and a decoder, both vanilla transformer models. The encoder only processes the visible patches and its output is passed through the decoder, which is typically much smaller than the encoder, to predict the values of the masked patches in the pixel space. In addition to being a highly efficient, generic, state-of-the-art SSL algorithm for video representation learning, MAEs also have the advantage of requiring very minimal data augmentation (we only use random resized crops and horizontal flips in the spatial domain). This is relevant for our purposes, because SSL algorithms that require heavy data augmentation strategies make the input less ``human-like''.

\begin{figure*}[t!]
  \centering
    \includegraphics[width=1.05\textwidth, clip]{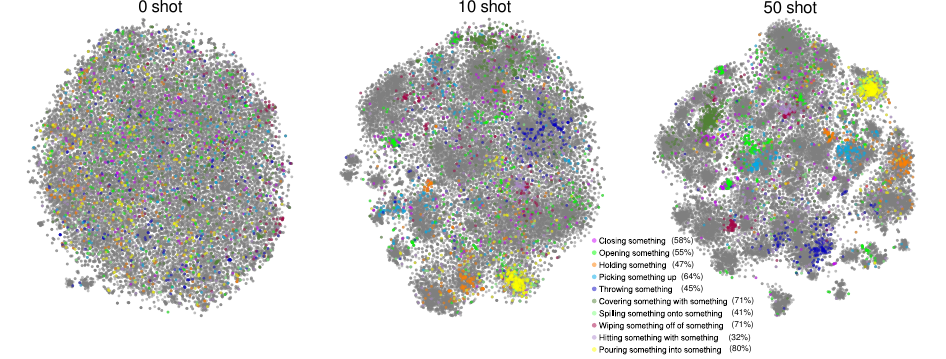}
    \vspace{-8mm}
  \caption{t-SNE embeddings of video clips from SSV2. Each point corresponds to a clip from the validation set.~Clips belonging to 10 ``developmentally realistic'' action categories (shown in the legend) are highlighted with different colors. (Left) Results from a 0-shot model pretrained on child S, but not finetuned with any data from SSV2. (Middle) Results from a model pretrained on child S and finetuned on 10-shot SSV2. (Right) Results from a model pretrained on child S and finetuned on 50-shot SSV2. Numbers in parentheses represent the top-5 validation accuracy for the corresponding categories in the 50-shot condition.}
  \label{embeddings_fig}
\end{figure*}

\subsection{Evaluation} Once the MAE models are pretrained with the child headcam data (or with other reference data), we evaluate the quality of the learned video representations through a number of downstream tasks. As is standard with MAEs, we use supervised finetuning in order to evaluate the models in downstream tasks. However, to ensure that the representations are learned mostly through SSL (and not through supervised finetuning) and also in the interest of psychological plausibility, we adopt a few-shot finetuning setting, where we only use a small number of labeled examples to finetune the models. 

For video-based evaluation, we consider two fine-grained action recognition tasks:~\textbf{Kinetics-700} and \textbf{Something-Something-V2} (SSV2) \cite{goyal2017}.~Kinetics-700 consists of short YouTube clips of 700 fine-grained action categories, whereas SSV2 contains short clips of people performing 174 different classes of fine-grained actions. The main difference between Kinetics-700 and SSV2 is that in SSV2, people perform \textit{instructed} actions, where the objects involved in the action are specified and carefully varied across clips to abstract the \textit{action concept} itself as a template. The videos in Kinetics-700, on the other hand, are ``in the wild'' clips downloaded from YouTube, so they are much less controlled with respect to potential shortcuts or biases in the data. 

For both tasks, we consider 10-shot and 50-shot supervised finetuning scenarios, using 10 and 50 training examples per class, respectively, to finetune the entire model (but with layer-wise learning rate decay as a regularizer so that top layers are modified more than bottom layers). In Kinetics-700, the 10-shot setting uses 17 hours of video and the 50-shot setting uses 87 hours of video in total for finetuning. In SSV2, the 10-shot setting uses roughly 2 hours of video and the 50-shot setting uses roughly 9 hours of video in total for finetuning. Thus, in every case, the amount of supervised video data used for finetuning is significantly less than the amount of video data used for SSL during pretraining (\textit{cf.}~194 hours of video from child S in SAYCam).

\section{Results}
\subsection{Result 1: \textit{After controlling for dataset size, video models trained with the child headcam data are competitive with models trained on a rich and diverse set of YouTube clips.}}

Figure~\ref{accuracies_fig}a-b show the results for the SSV2 and Kinetics action recognition tasks, respectively, in both 10-shot and 50-shot supervised finetuning conditions. We observe a substantial pretraining benefit in all cases: \textit{e.g.}~compare the models pretrained on {\color{orange}SAYCam} or {\color{cyan}Kinetics-700} with reference models trained from {\color{green}scratch} on the downstream task without any pretraining (Figure~\ref{accuracies_fig}). Models pretrained with the full Kinetics-700 dataset outperform all other models, but video pretraining with SAYCam also seems to be remarkably effective for downstream action recognition tasks. Comparing the model pretrained on child S with the model pretrained on the length-matched Kinetics-200h dataset, in particular, we find that they perform comparably in all cases, despite qualitative differences between the spatiotemporal characteristics of these two datasets.

Surprisingly, the model pretrained on the whole SAYCam dataset (SAY) does not perform better than the model pretrained on child S only. We conjecture that the presence of a number of low-quality, underexposed clips in recordings from the other two children (A and Y) may have affected the quality of the representations learned by the spatiotemporal MAE algorithm when trained with these data. We leave a more complete investigation of this finding to future work.

To investigate the extent to which these action recognition tasks genuinely require temporal information, as opposed to being susceptible to simpler image-based strategies such as recognizing the typical objects or scenes in individual frames that tend to co-occur with the given action categories, we next created a purely image-based reference model that did not use any temporal information (S-Img). This model was pretrained on the headcam data from child S using an image-based MAE \cite{he2022} with 80\% masking ratio and then finetuned on the downstream task, again in a purely image-based fashion (the model was evaluated by presenting the evaluation clips to the model frame by frame and averaging the predictions). This image-based reference model performed substantially worse than its video-based counterpart in SSV2 (\textit{cf.}~{\color{orange}S} \textit{vs.}~{\color{magenta}S-Img} in Figure~\ref{accuracies_fig}a), suggesting that this task genuinely requires learning the temporal regularities corresponding to different actions. However, the relative performance of S-Img was much better in the Kinetics-700 evaluation tasks (\textit{cf.}~{\color{orange}S} \textit{vs.}~{\color{magenta}S-Img} in Figure~\ref{accuracies_fig}b), suggesting that it may be possible to achieve high accuracies in this task by exploiting frame-based regularities and without using any temporal information at all. This result also illustrates the difficulty of creating datasets with ``in the wild'' internet data that are robust to shortcut learning strategies \cite{geirhos2020}.   

\subsection{Result 2: \textit{Video models trained with the child headcam data can learn to recognize developmentally realistic action categories from a small amount of labeled examples.}}

We next isolated 10 action categories in SSV2 that could be considered ``developmentally realistic'' in the sense that a 30-month old child would be expected to understand them (the cutoff age for SAYCam is 31 months). We manually selected these categories from SSV2 as follows: we only considered simple, general action categories that involve at most two objects, thus excluding more detailed action categories describing a specific mode of action, for instance, or more complex categories that involve more than two objects. For each of the remaining categories, we then checked the ``item trajectory'' of the basal verb describing the action on Wordbank \cite{frank2017} and included only those that are produced by most children by 30 months of age (\textit{i.e.}~``proportion of children producing'' $>$ 0.5). The final list of these developmentally realistic action categories can be found in the legend of Figure~\ref{embeddings_fig}.

Figure~\ref{embeddings_fig} shows t-SNE embeddings of these categories obtained from three different models: (i) a model pretrained on child S, but not finetuned on SSV2 at all (0-shot); (ii) a model pretrained on child S and finetuned on 10-shot SSV2; and (iii) a model pretrained on child S and finetuned on 50-shot SSV2. The 0-shot embeddings do not contain much category structure, attesting to the difficulty of this action recognition task. However, category-related semantic structure immediately begins to emerge as the model is finetuned with even a small amount of labeled examples (10-shot), and improves further with more labeled examples (50-shot).

Interestingly, we find that the model performs worse on this developmentally realistic subset of categories compared to the remaining categories: 56\% \textit{vs.}~64\% top-5 validation accuracy in the 50-shot condition (chance: 3\%). This could be because these are broader, more general categories compared to the rest. Indeed, among the categories with the highest accuracy in the 50-shot condition are: \textit{turning the camera left while filming something} (97\%), \textit{lifting a surface with something on it until it starts sliding down} (90\%), \textit{trying to pour something into something but missing so it spills next to it} (88\%), \textit{pouring something into something until it overflows} (88\%), \textit{pulling two ends of something so that it gets stretched} (87\%), \textit{twisting (wringing) something wet until water comes out} (83\%), and \textit{pretending to sprinkle air onto something} (81\%). These are typically more specific, more detailed action categories than the categories in our developmentally realistic subset.

\subsection{Result 3: \textit{Downstream performance scales favorably with the amount of child headcam data used for SSL.}}

\begin{figure}
  \centering
    \includegraphics[width=0.4925\textwidth, clip]{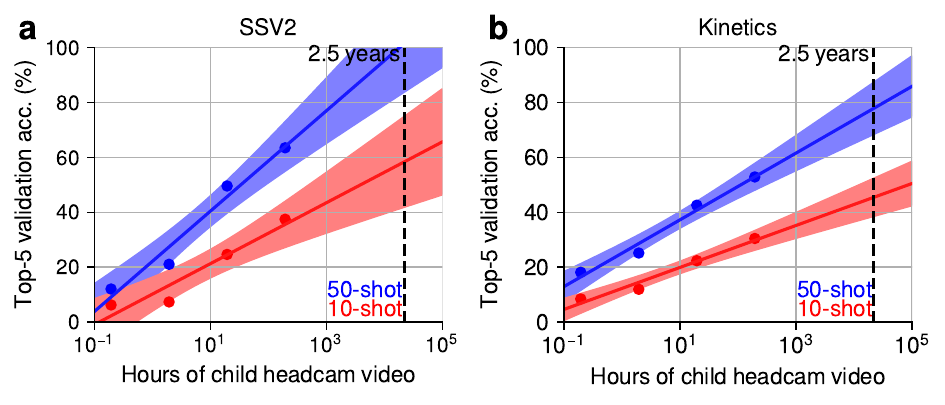}
    \vspace{-7mm}    
  \caption{A data size scaling experiment for child S. Spatiotemporal MAEs are trained on all data from child S and on subsets of it over a four orders of magnitude range in data size. The performance of the resulting models are evaluated in the SSV2 (a) and Kinetics-700 (b) action recognition tasks (individual dots). A log-linear model is fit to the data to extrapolate performance beyond the 194 hours of data we currently have from S. Shaded areas represent 95\% confidence intervals around the linear fits. A developmentally relevant time scale of 2.5 years is indicated by the vertical dashed line.}
  \label{scaling_fig}
\end{figure}

Although SAYCam is currently the largest developmentally realistic, longitudinal video dataset of its kind, the amount of data available from each child in SAYCam is quite limited compared to the amount of visual experience a child actually receives while growing up. For example, we have 194 hours of video available from child S, which corresponds to roughly 8 days (or equivalent to 16 days of visual experience if we factor in 12 hours/day of sleep). This is two orders of magnitude smaller than the amount of visual experience a child receives by the age of 4-5. This invites the natural question: what would happen if we had developmentally more realistic amounts of data, \textit{e.g.}~two orders of magnitude more data? 

To address this question, we performed a data size scaling experiment. We systematically varied the amount of video data from child S used for SSL pretraining: specifically, we trained spatiotemporal MAE models on all 194 hours of data (100\%), on 19.4 hours of data (10\%), on 1.94 hours of data (1\%), and finally on 0.194 hours of data (0.1\%) from S, thus covering a four orders of magnitude range in data size. We then evaluated these models on the downstream action recognition tasks and, using a simple log-linear scaling function, extrapolated their performance a few orders of magnitude beyond the amount of data we currently have from child S. 

The results of this scaling experiment are shown in Figure~\ref{scaling_fig}. We estimate substantial improvements in action recognition (on average, over $30\%$ absolute improvement in accuracy in the SSV2 50-shot task and $\sim20\%$ improvement in the Kinetics-700 50-shot task) with a two orders of magnitude increase in data size alone without any other changes. Further improvements may be possible by scaling up the input and model sizes as well \cite{orhan2023}.

\begin{figure*}[t!]
  \centering
    \includegraphics[width=1.0\textwidth, clip]{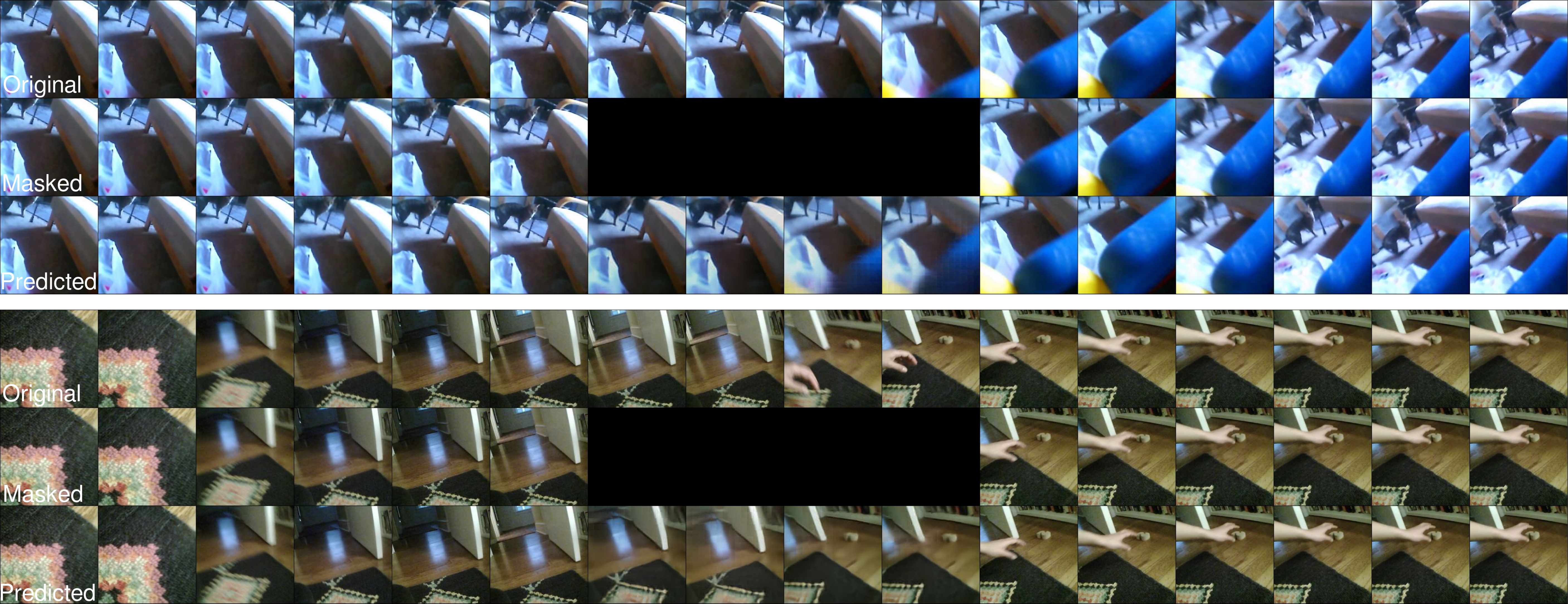}
    \vspace{-5mm}    
  \caption{Two examples illustrating the emergent video interpolation capabilities in a spatiotemporal MAE model trained on child S. In both cases, the top row shows the original sequence of frames; the middle row shows the masked sequence where the four central frames are masked out; and the bottom row shows the model completions of the masked frames. The original sequences are from the other two children in SAYCam (A and Y, respectively). Note that the model completions in both cases are not simple copies of the nearby visible frames; they are rather novel frames that indicate some understanding of the consequences of the camera movements on the image. Further examples can be found at \href{https://github.com/eminorhan/video-models/tree/master/comps}{this link}.}
  \label{interpolation_fig}
\end{figure*}

\subsection{Result 4: \textit{Emergent video interpolation capabilities in video models trained with the child headcam data.}}
We next devised a simple qualitative test to probe the models' understanding of the spatiotemporal dynamics over short video clips. In this test, we mask out the 4 central frames in the middle of a clip consisting of 16 consecutive frames. We then ask the model to ``fill in'' this central part, given the rest of the clip. To make sure that the behavior we observe is due to the out-of-distribution (\textit{ood}) generalization capacity of the model and not simply due to memorization or \textit{iid} generalization, we use the model trained on child S and evaluate it with clips from the other two children in SAYCam (A and Y). Note that this task is doubly challenging for the model: (i) the model was not trained with this particular masking strategy during pretraining (it was instead trained with random spatiotemporal masking as in Figure~\ref{maest_fig}) and (ii) the model did not see any of these clips during training. 

Despite these challenges, the model was often able to provide plausible completions of the clips. Visual inspection of the model's completions suggests that in many cases the model does not just utilize simplistic strategies such as copying the nearest visible frames, but rather generates plausible, novel completions that indicate some understanding of the movement of objects and other visual features across the image over the course of a clip. Figure~\ref{interpolation_fig} presents two qualitative examples illustrating this point.

\subsection{Result 5: \textit{Video models learn more accurate and more robust object representations than image-based models trained with the same data.}}
Finally, we sought to compare the \textit{object representations} that emerge in video models trained with the spatiotemporal MAE algorithm with those learned by image-based MAEs trained on the same data. We conjectured that since spatiotemporal MAEs learn to track visual representations over time, this could lead to more accurate or more robust object representations. To test this hypothesis, we evaluated both spatiotemporal MAEs and image-based MAEs trained with the headcam data from child S on two downstream visual object recognition tasks: ImageNet \cite{russakovsky2015} and out-of-distribution (OOD) ImageNet \cite{geirhos2021}. ImageNet is a standard benchmark for real-world visual object recognition and consists of a large collection of high-quality, photographic images belonging to 1000 different object categories. OOD ImageNet is an out-of-distribution version of ImageNet, which consists of 17 different evaluation tasks each generated by applying a different type of perturbation or transformation to images from the ImageNet validation set (\textit{e.g.}~changing the colors of the image, taking the silhouettes of the objects, stylizing the image, adding different types of noise to the image). Together, these two benchmarks evaluate the \textit{accuracy} and \textit{robustness} of real-word object recognition. 

We again adopt a few-shot supervised finetuning setup for our evaluations. Specifically, we finetune both video-based and image-based MAEs with 2\% of the training data from ImageNet, which corresponds to 25-26 images per class. To finetune the video model on image data, we simply repeat each image 16 times before feeding it to the model, thus effectively creating a static video clip with 16 frames. This allows us to use the pretrained video model without any alterations in the architecture.

\begin{figure}
  \centering
    \includegraphics[width=0.4625\textwidth, clip]{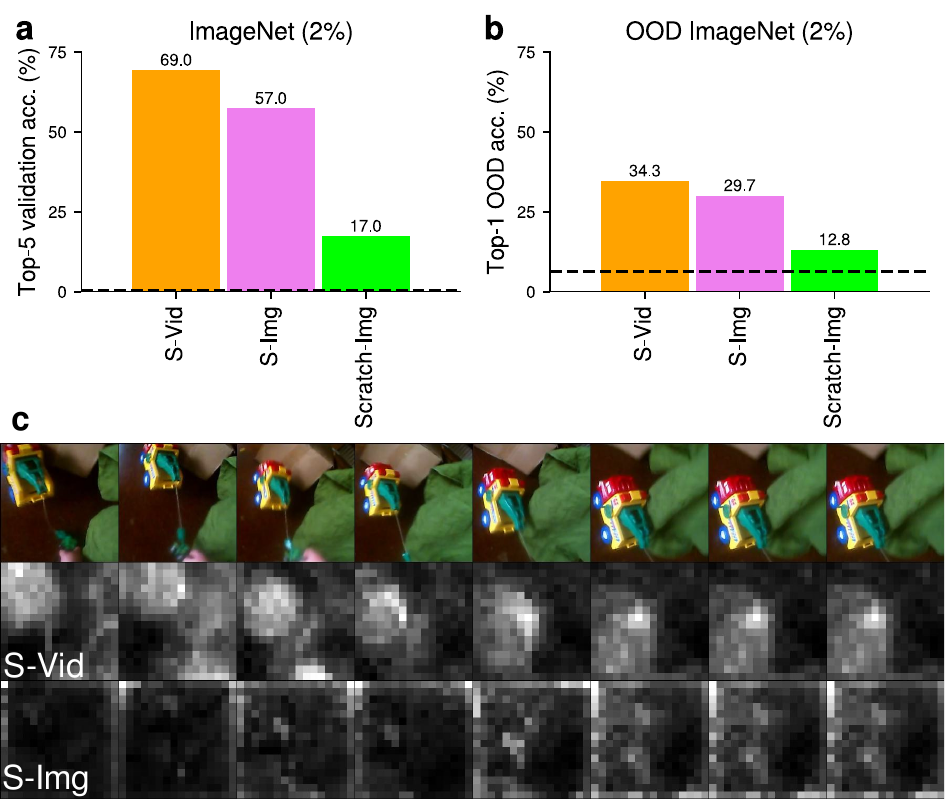}
    \vspace{-2.75mm}    
  \caption{Recognition accuracy on ImageNet (a) and on OOD ImageNet (b). {\color{orange}S-Vid} corresponds to a model pretrained with video-based SSL on child S; {\color{magenta}S-Img} is a model pretrained with image-based SSL on child S; {\color{green}Scratch-Img} is a model trained from scratch with 2\% of ImageNet training data without any pretraining. The pretrained models were also finetuned with 2\% of ImageNet training data. Dashed horizontal lines show the chance baseline.~(c) A simple sequence of frames from child S and the corresponding last-layer attention maps from {\color{orange}S-Vid} and {\color{magenta}S-Img}. Attention maps were obtained with respect to the \texttt{cls} token. Further examples can be found at \href{https://github.com/eminorhan/video-models/tree/master/atts}{this link}.}
  \label{objects_fig}
\end{figure}

The results are shown in Figure~\ref{objects_fig}a-b. On ImageNet, the model pretrained with video-based SSL on the headcam data from child S outperforms the model pretrained with image-based SSL on the same data ({\color{orange}S-Vid}: 69\% \textit{vs.}~{\color{magenta}S-Img}: 57\%). On OOD ImageNet, the video-based model again outperforms the image-based model ({\color{orange}S-Vid}: 34\% \textit{vs.}~{\color{magenta}S-Img}: 30\%), suggesting that the object representations learned by the video model are not only more accurate for fine-grained recognition, but also more robust to perturbations and transformations than those learned by the image-based model. The performance difference between the video model and the image-based model was particularly salient on the \textit{stylized} (+12\% in favor of the video model), \textit{sketch} (+9\%), and \textit{silhouette} (+4\%) subtasks in OOD ImageNet, suggesting a qualitatively different type of representation that may be more shape-sensitive and less texture-sensitive than in the image-based model. The respective attention maps from the two models seem to confirm this suggestion (Figure~\ref{objects_fig}c). We leave a more complete investigation of the possible reasons behind the superior object representations that seem to emerge with video-based SSL to future work (\textit{e.g.}~one possible factor is the difference in masking ratio per frame between image-based \textit{vs.}~video-based SSL: 80\% \textit{vs.}~90\%).

A recent work also reported more robust and more human-aligned object representations with video-based pretraining compared to image-based pretraining \cite{parthasarathy2023}. However, that work did not use the same datasets for video-based and image-based pretraining. In contrast, by using the exact same data (headcam videos from child S) and the same model architecture (ViT-H/14) for both video-based and image-based pretraining, we can isolate the effect of video-based \textit{vs.}~image-based pretraining objectives on the quality of the learned object representations.

\section{Discussion}
We trained video models on a large-scale, longitudinal dataset of headcam recordings collected from the perspective of a young child during their early development, using a highly generic SSL algorithm (spatiotemporal MAEs) and without assuming any strong inductive biases. These models learn a non-trivial amount of visual knowledge about temporal aspects of the world through self-supervised learning from a few hundred hours of headcam videos, enabling them to recognize challenging \textit{action concepts} from a small amount of labeled examples (Results 1-2) and to interpolate unseen videos with plausible and novel completions (Result 4). They also display favorable data size scaling in downstream tasks (Result 3), suggesting that we can expect to see substantial improvements in model capabilities if we can train our models with developmentally more realistic amounts of headcam data (\textit{i.e.}~roughly two orders of magnitude more data than we currently have). Finally, video models learn more accurate and more robust object representations that appear to be less texture-sensitive than image-based models trained on the same data (Result 5).

Our work has a number of limitations that should be kept in mind when interpreting our results. First, we only considered a limited number of tasks to evaluate the capabilities of the models. Future work should consider a larger, more diverse range of tasks that can probe the capabilities of the models, \textit{i.e.}~what they can and cannot do, in much more detail. In particular, intuitive physics is an important aspect of a child's internal model of the world that we have not rigorously evaluated in our models \cite{smith2019,piloto2022}. 

Second, although we were able to generate plausible completions with spatiotemporal MAEs (Result 4), MAEs, in general, are not designed to be generative models:~they do not have a well-defined likelihood function and, because of their mean squared error objective, they tend to generate low-quality, blurry predictions.~This suggests the need to train true generative video models like autoregressive models or diffusion models on SAYCam, in addition to MAEs. Such models would also enhance our understanding of model capabilities.

Third, throughout this work, we have adopted a few-shot supervised finetuning paradigm for evaluating our trained models (typically using only tens of labeled examples per class). However, it is unclear if this paradigm is psychologically realistic enough. Future work should aim to train models directly with developmentally realistic multimodal data sources instead, \textit{e.g.}~paired linguistic and visual inputs. This is crucial to more rigorously address the \textit{nature vs.~nurture} question that fundamentally motivates our work.

By demonstrating a sample of the non-trivial visual capabilities that can be learned generically from a few hundred hours of longitudinal headcam videos recorded from the perspective of a developing child, our work contributes to the burgeoning interaction between developmental psychology and modern machine learning \cite{zaadnoordijk}.

\bibliographystyle{apacite}

\setlength{\bibleftmargin}{.125in}
\setlength{\bibindent}{-\bibleftmargin}

\bibliography{cogsci,library_clean}

\end{document}